# Google Translate Error Analysis for Mental Healthcare Information: Evaluating Accuracy, Comprehensibility, and Implications for Multilingual Healthcare Communication

**Jaleh Delfani**
University of Surrey, UK
j.delfani@surrey.ac.uk

**Constantin Orăsan**
University of Surrey, UK
c.orasan@surrey.ac.uk

**Hadeel Saadany**
University of Surrey, UK
hadeel.saadany@surrey.ac.uk

**Özlem Temizöz**
University of Surrey, UK
o.temizoz@surrey.ac.uk

**Eleanor Taylor-Stilgoe**
University of Surrey, UK
e.j.taylor-stilgoe@surrey.ac.uk

**Diptesh Kanojia**
University of Surrey, UK
d.kanojia@surrey.ac.uk

**Sabine Braun**
University of Surrey, UK
s.braun@surrey.ac.uk

**Barbara Schouten**
University of Amsterdam, Netherlands
b.c.schouten@uva.nl

## Abstract

This study explores the use of Google Translate (GT) for translating mental healthcare (MHealth) information and evaluates its accuracy, comprehensibility, and implications for multilingual healthcare communication through analysing GT output in the MHealth domain from English to Persian, Arabic, Turkish, Romanian, and Spanish. Two datasets comprising MHealth information from the UK National Health Service website and information leaflets from The Royal College of Psychiatrists were used. Native speakers of the target languages manually assessed the GT translations, focusing on medical terminology accuracy, comprehensibility, and critical syntactic/semantic errors. GT output analysis revealed challenges in accurately translating medical terminology, particularly in Arabic, Romanian, and Persian. Fluency issues were prevalent across various languages, affecting comprehension, mainly in Arabic and Spanish. Critical errors arose in specific contexts, such as bullet-point formatting, specifically in Persian, Turkish, and Romanian. Although improvements are seen in longer-text translations, there remains a need to enhance accuracy in medical and mental health terminology and fluency, whilst also addressing formatting issues for a more seamless user experience. The findings highlight the need to use customised translation engines for Mhealth translation and the challenges when relying solely on machine-translated medical content, emphasising the crucial role of human reviewers in multilingual healthcare communication.

## 1 Introduction

The World Health Organisation (2019)[1] reported that around 970 million people worldwide, or 1 in 8 individuals, faced mental disorders, primarily anxiety and depressive disorders. The COVID-19 pandemic exacerbated these statistics, revealing a significant 26% increase in anxiety disorders and a notable 28% rise in major depressive disorders within a year (WHO, 2022). In the same vein, the Mental Health Foundation UK[2] highlighted that untreated mental health issues contribute to 13% of the global disease burden. Projections suggest that by 2030, mental health problems, especially depression, will become the leading cause of both mortality and morbidity globally.

---

[1] https://www.who.int/news-room/fact-sheets/detail/mental-disorders
[2] https://www.mentalhealth.org.uk/

Despite effective prevention and treatment options, a majority of those with mental disorders lack adequate access to care, especially true for migrants and refugees who may not speak the language of the country they are trying to settle in (Krystallidou *et al.*, 2024). In healthcare settings, obtaining human interpreters faces challenges such as waiting times, financial constraints, and limited availability (Al Shamsi *et al.*, 2020). In other instances where information is available in written format and translation, rather than interpretation, is needed, automated translation is occasionally employed. (Turner *et al.*, 2019; Chen and Acosta, 2016; Taylor-Stilgoe *et al.*, 2023).

Machine translation (MT) has emerged as a potentially valuable tool to overcome language barriers in healthcare, offering access to vital information for individuals with limited language proficiency. Generic MT tools like GT provide free access to automatic translation across many languages, but these translations vary in quality, raising concerns about reliability, liability, and data privacy, especially in safety-critical situations (Vieira *et al.*, 2021).

This paper explores the errors introduced by GT when used to access mental health-related materials such as website information and digital leaflets. It is organised as follows: Section 2 offers a brief literature review on the use of technology to facilitate communication in healthcare settings, particularly when participants lack a common language or when users need to comprehend a document written in an unfamiliar language. Section 3 outlines the methodology adopted in selecting data, translating it into other languages, and conducting an error analysis. Section 4 presents the findings related to two scenarios where GT was employed for data translation. The paper concludes with final remarks and suggestions for future research.

## 2  Literature Review

Effective communication in mental healthcare, especially through multilingual means, holds immense significance. Language serves as a channel for understanding, empathy, and successful treatment. Moreover, effectual multilingual communication dismantles cultural barriers, minimises stigma, and cultivates a sense of inclusivity. Studies indicate that migrants and refugees face an increased risk of developing depression and anxiety disorders due to exposure to stressors following resettlement, limited social support, and societal stigma and discrimination (Rousseau and Frounfelker, 2019). Furthermore, research provides evidence indicating higher prevalence rates of specific mental health disorders (*e.g.,* posttraumatic stress and psychosis-related disorders) among migrant populations compared to non-migrant populations (Priebe *et al.*, 2016). Language barriers can impede effective communication regarding treatment requirements and available care choices among patients, resulting in reduced utilisation of psychiatric healthcare services (Doğan *et al.,* 2019; Kiselev *et al.*, 2020; Krystallidou *et al.*, 2024; Marquine and Jimenez, 2020; Ohtani *et al.*, 2015). The presence of stigma and a hesitancy to seek assistance further exacerbates the language barrier (Giacco *et al.*, 2014). Overcoming language barriers is essential for achieving high levels of satisfaction among both medical professionals and patients, ensuring proper treatment, and maintaining patient safety (Al Shamsi *et al.*, 2020). There are two primary solutions to address language barriers: utilising interpreting services and leveraging available translation applications.

Interpreting services may escalate both the cost and duration of the treatment process (Al Shamsi *et al.*, 2020) and might not consistently be accessible (Arafat, 2016; Doğan *et al.*, 2019; Felsman *et al.*, 2019; Khanom *et al.*, 2021; Pallaveshi *et al.*, 2017; Shrestha-Ranjit *et al.*, 2017). In such scenarios, machine translation emerges as the most readily available solution.

Generic tools like GT or bespoke solutions (Dew *et al.*, 2018; Haddow *et al.*, 2021; Vieira *et al.*, 2021) play a crucial role in easing communication between healthcare professionals and patients who lack a shared language. In such scenarios, direct interaction occurs between the patient and healthcare provider involving a device and interface granting access to machine translation software. This dynamic interaction takes place in co-located settings where patients and providers use smartphones or tablets to access translation software. Alternatively, it occurs in situations where they are connected remotely through communication technology, as exemplified by telehealth consultations employing integrated translation tools such as Skype Translator[3].

Apart from being insufficiently explored concerning the complex interactions within the use of machine translation tools in interpersonal healthcare communication, a significant obstacle to their practical implementation is the issue of quality. Currently, machine translation fails to provide accurate mediation for numerous language pairs in diverse healthcare settings. To tackle the existing challenges associated with MT while still incorporating a degree of automation, various semi-automated approaches, particularly phrase-based translation apps such as Xpromt and BabelDr, have been developed (Braun *et al.*, 2023). These apps are typically pre-loaded with validated human translations of common phrases and sentences, providing essential communication support in specific healthcare settings. The interaction with these apps can be as intricate as the interaction with pure MT software.

Over the past decade, publicly available generic machine translation tools have shown improvement. Translation applications like GT and Microsoft Translator now provide the translation of written and spoken input into text and/or speech output in near-real time for an expanding range of language pairs. The utilisation of generic MT tools in daily clinical practice became apparent in a study investigating attitudes toward vaccination among Polish and Romanian communities in England. A significant number of healthcare professionals delivering vaccines to these communities reported relying on free MT tools to communicate with people who did not speak English (Moberly, 2018a). Although official guidance in the UK does not endorse their use in medical consultations, healthcare workers perceived these tools as more accessible than professional interpreting services, especially in time-pressured appointments. In response to this discovery, medical advisers emphasised the potential risks of using tools like GT in everyday clinical practice, citing the possibility of introducing communication errors and compromising patient safety, which could expose doctors to legal action (Moberly, 2018b). However, the advisers acknowledged that MT tools might have a limited role in emergencies or other exceptional circumstances.

GT stands out as one of the most recognised and extensively utilised machine translation tools among the general public. Supporting translations among 133 languages (as of May 2022)[4] and compatible with both iOS and Android systems, this free application is a popular choice. Platforms like X (former Twitter) often rely on GT to offer translation services, and users are well-acquainted with its functionality. According to reports from 2021, the tool translates over 100 billion words daily[5], indicating that the public is inclined to use it for their translation needs.

---

[3] https://www.skype.com/en/features/skype-translator/
[4] https://blog.google/products/translate/24-new-languages/
[5] https://ttcwetranslate.com/how-does-google-translate-work/

Research indicates that refugees frequently employ GT on their smartphones as their primary online translation tool (Abujarour, 2022) and it serves as the most accessible and free primary means of communication in healthcare settings where language is a barrier. Nevertheless, it is crucial to grasp the accuracy and potential drawbacks of GT output, especially when dealing with sensitive and critical healthcare content (Leite *et al.*, 2016).

In evaluations of GT's effectiveness in translating emergency department discharge instructions from English to Spanish and Chinese, Khoong *et al.* (2019) discovered that a substantial percentage of sentences were accurately translated (92% for Spanish and 81% for Chinese). However, they noted that 2% of Spanish and 8% of Chinese sentence translations had the potential for significant or life-threatening harm, primarily due to errors in word disambiguation. In a parallel study examining additional language pairs, Taira *et al.* (2021) observed that GT output was inconsistent across six language pairs, with accuracy rates ranging from 55% to 94%. Assessing another generic translation app, iTranslate, in translating common questions posed by diabetes patients to clinicians, Chen, Acosta, and Barry (2017) found that the MT output was comparable to human translation in terms of accuracy for simple sentences but error-prone for complex sentences.

The National Health Service (NHS) in England explicitly advises its staff against utilising online MT services, citing the lack of assurance regarding the quality of translations (NHS England, 2023). Nevertheless, instances are prevalent where healthcare staff resort to non-specialised, commercially available MT tools, such as GT, when providing interpersonal or written assistance to patients with limited to no proficiency in the English language (Bell *et al.*, 2020; Moberly, 2018a; Royal College of Midwives, 2017). Vieira *et al.*, (2021) highlight that research on the implications of the widespread and potentially uninformed use of this technology remains limited. Studies investigating the impact of MT on patient medical record documentation reveal that healthcare professionals are largely unaware of the errors that MT can introduce, particularly concerning abbreviations (Taylor-Stilgoe *et al.*, 2023).

In light of these concerns and knowledge gaps, the objective of this investigation is to explore the inaccuracies introduced by GT when translating materials related to mental health from English to five languages, each with differing levels of resources. By scrutinising the accuracy and potential pitfalls of MT output in this critical healthcare context, this research seeks to contribute insights into the nuanced challenges and opportunities associated with the use of MT tools in mental health communication.

## 3 Methodology

This section outlines the methodology employed in this paper for the preparation of the investigated datasets, the utilisation of the machine translation engine, and the assessment of translation quality.

### 3.1 Datasets

For our study, we utilised two datasets comprising sentences and documents written in English that contain information related to mental health. The first set of sentences was extracted from the UK NHS website[6], which provides healthcare information to patients. This website was chosen for its comprehensive resource on health conditions, symptoms, and treatments. It features a guide crafted by healthcare professionals that offers insights into a variety of health issues, advising visitors on what actions to take and when to seek assistance. From these articles, we extracted 100 English sentences (1494 words) related to the mental health domain, which were then translated into other languages using GT. We will refer to this dataset as the "NHS dataset".

---
[6] https://www.nhs.uk/conditions/

The second dataset was constructed using digital information leaflets sourced from the UK Royal College of Psychiatrists[7]. We will refer to this dataset as the "RCP dataset". These leaflets are originally written in English and present user-friendly and evidence-based information on mental health problems, treatments, and related subjects. Qualified psychiatrists, with input from patients and carers, contribute to the creation of these informative materials. For our experiments, we selected the leaflet with the topic of "Depression" (1267 words). We used GT to translate our datasets into five languages under investigation in this study, namely Persian, Modern Standard Arabic, Turkish (low-resourced languages), Romanian (a medium-resourced language), and Spanish (a high-resourced language). Our objective was to evaluate GT's performance in the mental health context across languages with varying levels of resources in two scenarios: a) translating individual sentences and b) translating longer texts (contextualised paragraphs).

### 3.2 Data Preparation

The NHS dataset was translated into the aforementioned languages using GT and organised into separate spreadsheets. Subsequently, native speakers of each respective language conducted manual analyses (the analysis procedure will be elaborated upon in the next section). For the second dataset, which comprises digital leaflets, we opted to provide context to GT to assess whether its performance differed from the first scenario where individual sentences were translated. To be more specific, in our study, GT was stress-tested on two types of data: a) individual sentences which may lack the overall context; b) longer stretches of text from leaflets as contextually coherent units.

Throughout all the experiments detailed in this paper, we employed the online version of GT, without any customisation or tuning for a specific domain. The translations were conducted using the version available in June 2023.

### 3.3 Evaluation Method

Assessing the output of machine translation poses a challenging task that has undergone extensive examination. Commonly employed methods for evaluation include automatic approaches such as BLEU (Papineni *et al.*, 2002) and METEOR (Banerjee and Lavie, 2005), which automatically compare machine-generated output with a reference translation. While these methods prove valuable in the development of machine translation systems, they fail to elucidate why a translation falls short. Consequently, we opted for manual error analysis. Despite its time-consuming nature, this approach proves invaluable in discerning the specific errors made by MT engines. Following a preliminary analysis of MT output on our datasets, we developed an error taxonomy to systematically capture the most significant errors in the mental health context.

---

[7] https://www.rcpsych.ac.uk/mental-health

Our error typology comprised the following:

- **Inaccuracy of mental health and medical terminologies:**
  Instances where either or both of these aspects have not been translated accurately, and therefore, can have consequences for the effectiveness of the message. To ensure consistency and reliability in identifying such terms, the sentences in our datasets have been reviewed by cross-referencing them with several resources, including the World Health Organisation key terms and definitions in mental health[8], NHS mental health conditions[9], and Bupa mental health glossary[10]. It should be mentioned that out of 100 sentences in our NHS dataset, 53 included such terms.
- **Syntactic/semantic errors:**
  These errors may arise from incorrectly constructed target sentences or inaccuracies in translating words or phrases.
- **Comprehensibility issues:**
  Translations that are intricate and challenging for individuals with diverse levels of mental health literacy to understand.
- **Fluency issues:**
  Translations that affect the natural flow and readability of the content.
- **Clarity and Coherence issues:**
  Instances where translations do not read smoothly and coherently in the target language and cause ambiguity.
- **Culturally insensitive translations:**
  Renderings that are deemed unacceptable due to the cultural nuances of the target language. This type of error is especially crucial, considering the divergent perspectives on mental health prevalent across various cultures.
- **Critical errors:**
  These are errors that significantly impact the meaning of the message and may pose a risk to patient health outcomes.

To conduct our error analysis, native speakers of the target languages were recruited to review the output of machine translation along with the source texts in English. They were tasked with identifying translation errors and categorising them into the aforementioned classes. The evaluators had a linguistic background and were briefed on the error classification guidelines. It should be noted that the analysis presented in this paper concentrated on the linguistic aspects of the translation and the readers' comprehension of the message, rather than attempting to evaluate the quality of the translation from a healthcare perspective. Such an evaluation is slated for future consideration.

## 4   Findings

The results will be presented in two parts: the first part will discuss the findings collectively, i.e., based on the phenomena we observed across languages, while the second part will showcase more detailed findings with examples related to each language under investigation.

---

[8] https://www.who.int/southeastasia/health-topics/mental-health/key-terms-and-definitions-in-mental-health
[9] https://www.nhs.uk/mental-health/conditions/
[10] https://www.bupa.co.uk/~/media/files/mms/bins-02812.pdf

**Collective Results for the NHS Dataset:**

After analysing the GT output for 100 sentences from the NHS website, it was observed that Arabic exhibited the highest error rate in almost all categories among the other languages under investigation (Figure 1). This was followed by Persian and Romanian, respectively. The quality of the Spanish translation was not as good as that of Turkish. Turkish had the lowest number of errors among the languages we investigated in this dataset.

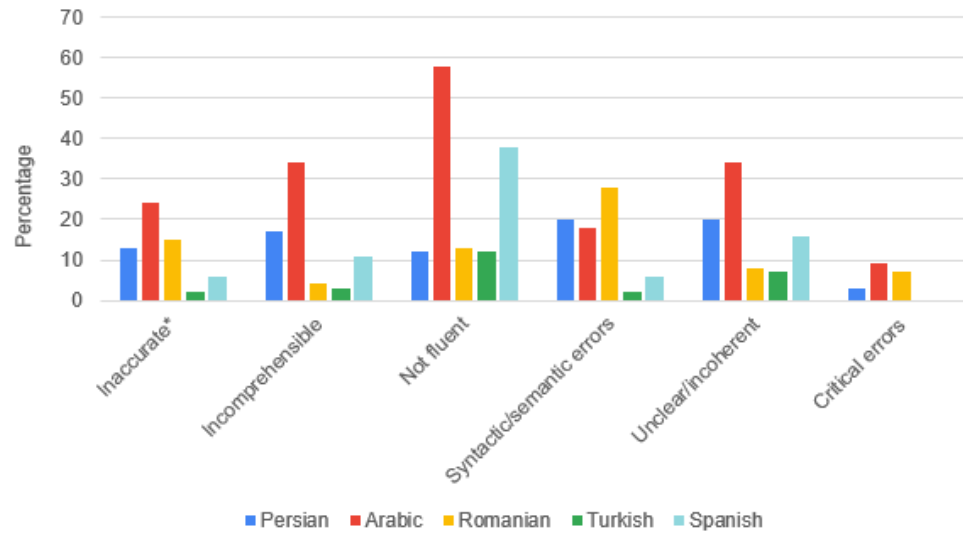

Figure 1: Relative frequency of the error types in all languages- NHS dataset

As seen in Table 1 below, Arabic exhibited the highest number of errors in each category (shown in red), except for the syntactic/semantic errors, where Romanian had the highest error rate. Seven of these errors in Romanian could pose a risk to the patient's life. Critical errors were also observed in Arabic and Persian, with Arabic having the highest number of this error type. Fluency issues were among the most frequently occurring errors in almost all languages, even in Spanish, which is considered a high-resourced language. Turkish demonstrated the lowest error rate in almost all categories among other languages (shown in green), which is rather surprising, as this language falls into the same category of low-resourced languages as Arabic and Persian. No cases of cultural insensitivity were observed in GT output for this dataset.

| Error Type | Persian | Arabic | Romanian | Turkish | Spanish |
|---|---|---|---|---|---|
| Inaccurate terminology | 13 | 24 | 15 | 2 | 6 |
| Incomprehensible | 17 | 34 | 4 | 3 | 11 |
| Not fluent | 12 | 58 | 13 | 12 | 38 |
| Syntactic/semantic errors | 20 | 18 | 28 | 2 | 6 |
| Incoherent | 20 | 34 | 8 | 7 | 16 |
| Culturally insensitive | 0 | 0 | 0 | 0 | 0 |
| Critical errors | 3 | 9 | 7 | 0 | 0 |

Table 1: Absolute frequency of the error types in all languages- NHS dataset

**Collective Results for the RCP Dataset:**

For this particular dataset, we opted for a contextual analysis, focusing on paragraphs rather than individual sentences, a method distinct from our approach with the other dataset. This involved the utilisation of a Word document, where we systematically examined each paragraph, annotating observed errors. Subsequently, evaluators documented the prevalent error types in each language, and where possible, identified potential causes for these errors.

The findings demonstrated a notable divergence from those obtained with the NHS dataset. Arabic exhibited superior translation quality, with only minor pronoun errors detected. In contrast, Persian presented challenges in punctuation, bullet point formatting, and code-switching in the GT output, leading to syntactic/semantic errors, incomprehensibility, and coherence issues. Both Romanian and Turkish exhibited issues stemming from bullet point structure, contributing to syntactic and semantic errors, comprehensibility problems, and fluency issues. It was unexpected to observe a decline in GT output quality for Turkish when translating paragraphs compared to individual sentences in the other dataset. This decline was mainly caused by the bullet-pointed structure of the source text i.e., the English information leaflets. Spanish, in this dataset, encountered problems related to comprehensibility, fluency, coherence, lack of gender agreement, and incorrect/missing abbreviations.

**Outcomes Specific to Each Language:**

**Persian**

**NHS dataset:**

In the analysis of the NHS data, 53 sentences contained medical or mental health terminology. Persian translations showed 13% inaccuracies in such terms, with 3 cases rendering messages incomprehensible and 2 affecting overall fluency. Of 100 Persian translations, 17 were incomprehensible, 12 had fluency issues, and 20 exhibited syntactic/semantic errors, 3 of which were critical. Additionally, 20 incoherent sentences were documented.

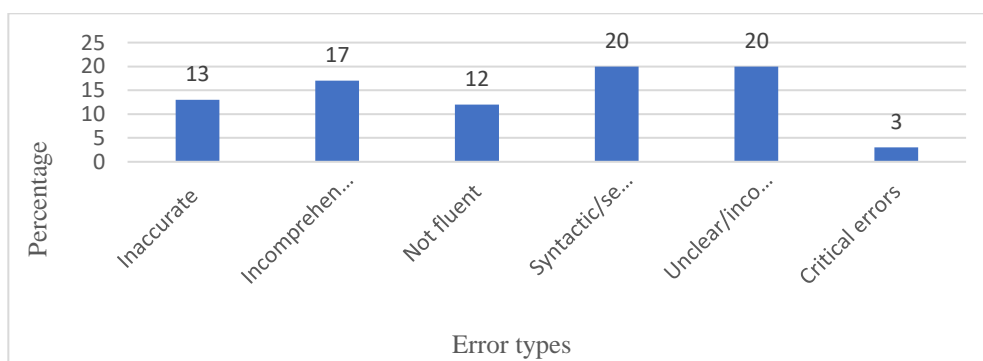

Figure 2: GT error analysis for Persian

The following are illustrative examples of the above-mentioned errors identified within this dataset.

| Error type | Source text | GT output | Back translation |
|---|---|---|---|
| Comprehensibility | Give a fractional pause after each expiration and inspiration. | پس از هر انقضا و الهام، مکث کسری بدهید. | After each expiry and inspiration, give a fractional pause. |
| Critical | The diet of persons suffering from depression should completely exclude tea, coffee, alcohol, chocolate and cola, all white flour products, sugar, food colourings, chemical additives, white rice and strong condiments. | رژیم غذایی افراد مبتلا به افسردگی باید به طور کامل شامل چای، قهوه، الکل، شکلات و کولا، تمام محصولات آرد سفید، شکر، رنگ های غذایی، افزودنی های شیمیایی، برنج سفید و چاشنی های قوی باشد. | The diet of persons suffering from depression should completely include tea, coffee, alcohol, chocolate and cola, all white flour products, sugar, food colourings, chemical additives, white rice and strong condiments. |

Table 2: Examples of errors for GT output in Persian- NHS dataset

In the first example provided, the terms 'expiration and inspiration' are translated as 'the expiry date' and 'being inspired to do something', rather than 'exhale and inhale' in this context. This interpretation can certainly result in a misunderstanding of the original message, impacting its comprehensibility and clarity. In the second example above, falling into the category of critical error cases, the term 'exclude' has been translated to its complete opposite, 'include', which can pose a risk to patient safety and health outcome.

**RCP dataset:**
The main challenges faced by GT in translating Persian mental health-related leaflets were related to two key issues: problems with punctuation, particularly in translating bullet point formatting, and the occurrence of code-switching between Persian and Latin scripts in the translated content. These challenges primarily led to syntactic errors, resulting in a significant loss of comprehensibility and clarity in the translated text, as well as compromised linguistic fluency. The analysis identified six critical errors in the dataset, highlighting the need for improvement in handling these specific issues. Table 3 below illustrates one of such errors.

| Error type | Source text | GT output | Back translation |
|---|---|---|---|
| Critical | In your mind, you lose your self-confidence, start to feel hopeless, and perhaps even suicidal. | در ذهن شما، شما اعتماد به نفس خود را از دست بده شروع به احساس ناامیدی و شاید حتی خودکشی کن. | In your mind, lose your confidence, start to feel hopeless, and even kill yourself. |

Table 3: Example of errors for GT output in Persian- RCP dataset

In the example mentioned above, a critical error is observed in the translation. The original sentence, which provides information regarding the symptoms of depression, has been translated in an imperative sense, urging the patient to lose their confidence, start feeling disappointed, and maybe even commit suicide. This is an extremely serious error that could potentially lead to the patient contemplating self-harm or suicide. Such critical errors pose a significant risk to the well-being and safety of the patients.

## Arabic

**NHS dataset:**

In the English to Arabic translation, 24% of the 53 medical terms analysed were mistranslated, a higher rate than in Persian. The lack of fluency in the Arabic dataset was notably high at 58%, and 34% of translated sentences were incomprehensible. Arabic translations exhibited a relatively higher number of critical errors, where syntactically correct sentences provided incorrect information in the target language. Examples include the translation of 'mantras' as singing, leading to a loss of essential mental health advice, and the reversal of advice from "practice yoga and meditation" to "avoid yoga and meditation," creating challenges in detecting errors due to the fluency and syntactic correctness of the Arabic sentences (Table 4).

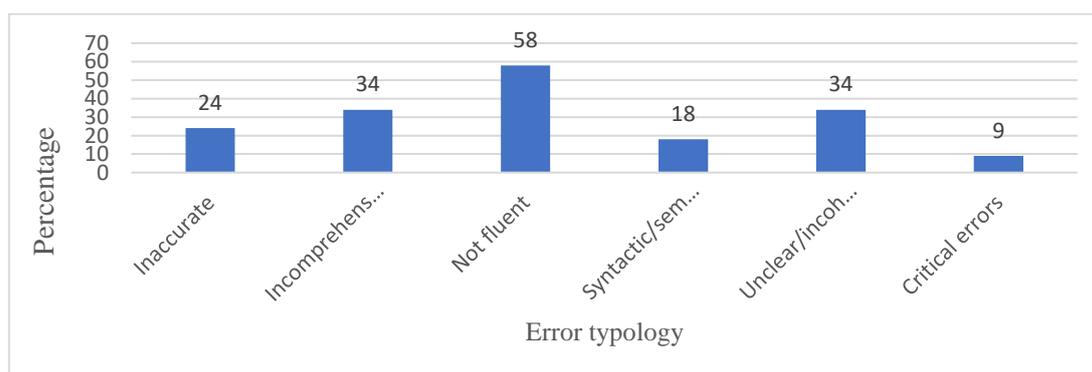

Figure 3: GT error analysis for Arabic

| Error type | Source text | GT output | Back translation |
|---|---|---|---|
| Semantic | Focus your mind on mantras or breathe. | ركز عقلك على التغني أو التنفس. | Focus your mind on singing and breathing |
| Critical | Practice yoga or meditation to avoid stress in life. | تجنب ممارسة اليوجا أو التأمل ضغط في الحياة. | Avoid yoga or meditation on stresses of life. |

Table 4: Examples of errors for GT output in Arabic- NHS dataset

**RCP dataset:**
Unlike the translation of NHS sentences, English to Arabic translation of longer medical leaflets showed a higher standard. The Arabic translation of the depression leaflet was fluent and comprehensible, with minor errors involving pronoun choices. Overall, the performance of GT was notably better with longer text spans in the translation from English to Arabic.

## Turkish

**NHS dataset:**

GT's Turkish output resulted in high-quality output with no critical errors. The most common issues were related to fluency, accounting for 12% of the translation output. However, these fluency problems did not impact the overall clarity or comprehensibility of the content. Additionally, 7% of the translated sentences were deemed unclear or incoherent, while 3% were incomprehensible. Only 2% of sentences containing mental health terminology were inaccurately translated and these were not considered critical.

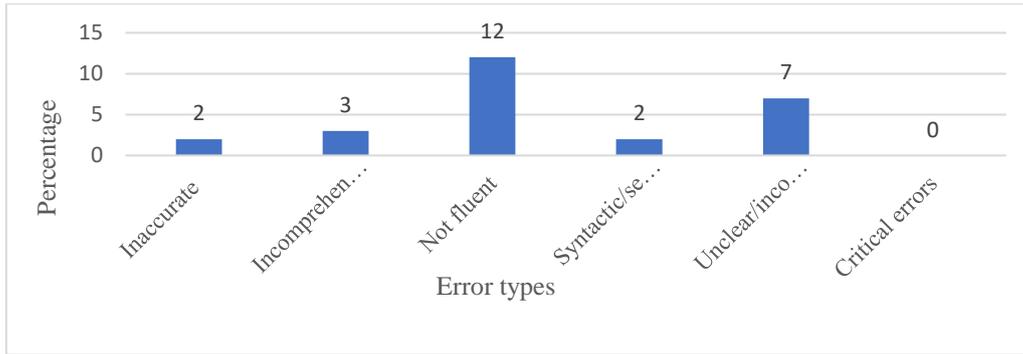

Figure 4: GT error analysis for Turkish

| Error type | Source text | GT output | Back translation |
|---|---|---|---|
| Semantic | Resist the temptation to drown your sorrows with alcohol. | alkolle Acılarınızı cazibesine boğmanın direnin. | Resist the temptation to drown your sorrows with alcohol. |

Table 5: Example of errors for GT output in Turkish- NHS dataset

In the above example, the infinitive 'to drown' is literally translated as 'boğmanın' which would be appropriate 'to drown someone in water'. However, in the given context, a fluent translation would require the use of 'Acılarınızı alkolle bastırmanın cazibesine direnin' (resist the temptation to suppress your sorrows with alcohol).

**RCP dataset:**

For this dataset, GT yielded lower quality in Turkish translation compared to the NHS dataset. While there are no critical errors, the bullet-point structure led to fluency issues, affecting readability. Due to the bullet-pointed structure, similar to the GT output in Persian, the information regarding the symptoms of depression in the original text has been translated in an imperative sense. The translation urges the patient to lose their confidence, start feeling hopeless, and maybe even become suicidal. This raises concerns about potential risks in the translation pattern of turning bullet points into imperatives. Medical and mental health-related terminologies have been translated accurately to Turkish; however, an inconsistency was observed related to the rendition of medical acronyms in this dataset. For example, the phrases such as 'cognitive behavioural therapy (CBT)' and 'selective serotonin reuptake inhibitor (SSRI)' are translated correctly as 'bilişsel davranışçı terapi (CBT)' and 'seçici serotonin geri alım inhibitörü (SSRI)', respectively; however, the acronyms are left untranslated. On the other hand, in one instance, the acronym is translated correctly when it is used without the expanded version, i.e., 'CBT programmes' as 'BDT programları'. In another one, however, both the therapy name and the acronym were correctly translated; for example, Electroconvulsive therapy (ECT) is translated as Elektrokonvülsif tedavi (EKT).

**Romanian**

**NHS dataset:**
The predominant error type in GT output for Romanian was syntactic/semantic errors, constituting 28 percent of the errors, with 7% classified as critical errors. For example, in Table 6 below, the word "high", translated as "big" in Romanian, fails to convey that the source sentence discusses being high due to drugs. Out of a total of 53 sentences containing medical/mental health terminology, 7 of them (13%) misinterpreted one or more terms within the sentence. Nevertheless, in most cases, while the translations may seem peculiar, they are likely to provide the reader with an understanding of the intended meaning, and therefore, they are not deemed critical errors.

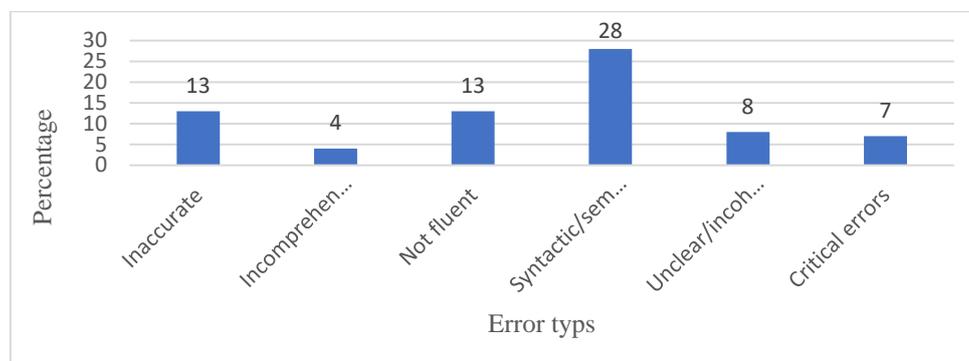

Figure 5: GT error analysis for Romanian

| Error type | Source text | GT output | Back translation |
|---|---|---|---|
| Semantic | You may also have to encourage the person to visit the psychiatrist or ask for urgent help whenever the sufferer is so "high" that he or she is no longer aware that anything is wrong. | De asemenea, este posibil să fiți nevoit să încurajați persoana să viziteze psihiatrul sau să cereți ajutor urgent ori de câte ori suferința este atât de "mare", încât nu mai este conștient de faptul că ceva nu este în regulă. | In addition, it is possible that you have to encourage the person to visit the psychiatrist or to ask for urgent help whenever the suffering is so big that he is no longer aware that something is not right. |

Table 6: Examples of errors for GT output in Romanian-NHS dataset

**RCP dataset:**
The analysis of the leaflets revealed numerous errors in GT Romanian output, mainly due to bullet point formatting. Errors included verb/pronoun disagreements, incorrect verb forms, and distorted sentences. The formatting issue led to critical errors, especially in translating from second person to third person or infinitive forms. For instance, the translation of "[you] can't eat and lose weight" (as symptoms of depression) results in an inaccurate message: "they can't eat and cannot lose weight". Moreover, translation errors can occasionally cause a shift in focus from the reader to a broader audience, causing confusion. These issues are primarily linked to bullet point formatting, as paragraphs without special formatting were translated with fewer mistakes.

**Spanish**

**NHS dataset:**
For the English to Spanish output, 3 of 53 such instances were considered to have been rendered inaccurately, accounting for 6% of the corresponding translations. In 2 of the 3 cases, the inaccuracy of the medical terminology negatively impacted on the comprehensibility of the intended message, though none were observed as containing critical errors. While 38 out of 100 sentences were observed as containing instances of disfluency, only 6 contained semantic or syntactic errors that could impact on the target reader's understanding of the mental health message.

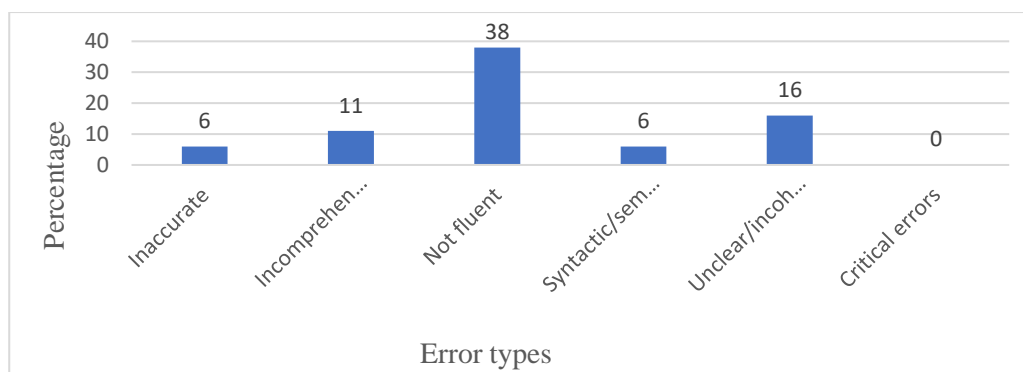

Figure 6: GT error analysis for Spanish

| Error type | Source text | GT output | Back translation |
|---|---|---|---|
| Semantic | Sleep is disturbed when you have a maladjusted family. | El sueño se molesta cuando tienes una familia desadyectada. | Sleep is disturbed when you have a [incomprehensible] family. |

Table 7: Example of errors for GT output in Spanish- NHS dataset

In the given passage, the adjective 'desadyectada' is highlighted as a significant impediment to the quality of translation. This term, used to render 'maladjusted', not only fails to convey the original meaning but also appears to be entirely invented by GT.

**RCP dataset:**

The translation of medical leaflets from English to Spanish faced significant challenges, including inconsistencies in subject pronouns and verb conjugations. Issues included mismatches between informal and formal 'you', incorrect verb forms, and inappropriate word choices. Gender agreement problems, incorrect abbreviations, untranslated terms, and missing articles were also noted. However, the text quality improved when bulleted formatting was reduced.

## 5   Concluding Remarks

Our research underscores the critical importance of recognising and addressing the limitations inherent in the use of Google Translate within the context of mental health. It is crucial to adopt a cautious approach and implement necessary precautions to safeguard patient well-being and facilitate effective communication. One prominent aspect that emerges from our findings is the pressing need for substantial enhancements in GT's performance, particularly within the realm of mental health. This necessity becomes even more pronounced when considering languages with limited resources such as Arabic and Persian, where GT may exhibit shortcomings that could impede communication and understanding and potentially pose a risk to patient well-being and safety. Recognising and addressing these issues promptly and advocating for improvements in the accuracy of translations is significant, especially in areas that involve intricate structures, such as bullet points, code-switching, and specialised medical and mental health terminologies.

Furthermore, it is crucial to acknowledge that GT is susceptible to errors at any given point, even within high-resourced languages, and should not be the sole reliance in sensitive contexts such as mental health. Recognising and embracing the collaborative role of human reviewers is integral to the responsible and effective use of machine translation tools like GT in the mental health context.

## 6  Future Work

Our research is ongoing and is part of a broader research initiative. Possible areas for improving this research include:

- Broadening the scope of our language coverage and augmenting the sample size. By encompassing a more extensive array of languages and increasing the number of evaluators in our study, we anticipate obtaining a more comprehensive understanding of the nuances involved.
- Undertaking a comparative analysis, for instance, comparing the performance of GT versus ChatGPT. This comparative approach will allow us to determine strengths, weaknesses, and potential areas for refinement in this context.
- Conducting case studies, involving real users in specific scenarios and practical applications that may not be apparent through quantitative analysis alone, can be explored to gain further insights.
- Examining the output generated by GT within the mental healthcare domain, taking the mental healthcare professionals' perspective in the analysis process. This can be done to assess whether the identified errors have the potential to impact patient health outcomes, shifting the focus beyond linguistic elements alone.

**Acknowledgments**

This project was funded by the European Union Asylum, Migration and Integration Fund [Award No:101038491]. The views and opinions expressed in this paper are solely those of the authors and do not necessarily represent the official views of the funding agency. The funding agency is not responsible for the ideas expressed or conclusions drawn in this manuscript.